\documentclass[journal]{IEEEtran}
\usepackage{graphicx}
\usepackage{amsmath}
\usepackage{amssymb}
\usepackage{subcaption}
\usepackage{booktabs}
\usepackage{float}
\usepackage{multirow}
\usepackage[ruled,vlined]{algorithm2e}
\usepackage{titlesec}
\usepackage{caption}
\titlespacing{\section}{0pt}{5pt}{5pt}
\titlespacing{\subsection}{0pt}{4pt}{4pt}
\titlespacing{\subsubsection}{0pt}{4pt}{4pt}

\ifCLASSINFOpdf
\else
\fi
\begin{document}

\title{Dynamic Game-Theoretical Decision-Making Framework for Vehicle-Pedestrian Interaction with Human Bounded Rationality}
%
%
%

\author{Meiting Dang,
        Dezong Zhao,~\IEEEmembership{Senior Member,~IEEE,}
        Yafei Wang,~\IEEEmembership{Member,~IEEE}
        and Chongfeng Wei*,~\IEEEmembership{Member,~IEEE}
        
\thanks{*Chongfeng Wei is the corresponding author.}    
\thanks{Meiting Dang, Dezong Zhao and Chongfeng Wei are with James Watt School of Engineering, University of Glasgow, Glasgow, G12 8QQ, United Kingdom (email: m.dang.1@research.gla.ac.uk, dezong.zhao@glasgow.ac.uk, chongfeng.wei@glasgow.ac.uk)}
\thanks{Yafei Wang is with the School of Mechanical Engineering, Shanghai Jiao Tong University, Shanghai 200240, China (email: wyfjlu@sjtu.edu.cn)}}


%

\maketitle

\begin{abstract}
Human-involved interactive environments pose significant challenges for autonomous vehicle decision-making processes due to the complexity and uncertainty of human behavior. It is crucial to develop an explainable and trustworthy decision-making system for autonomous vehicles interacting with pedestrians. Previous studies often used traditional game theory to describe interactions for its interpretability. However, it assumes complete human rationality and unlimited reasoning abilities, which is unrealistic. To solve this limitation and improve model accuracy, this paper proposes a novel framework that integrates the partially observable markov decision process with behavioral game theory to dynamically model AV-pedestrian interactions at the unsignalized intersection. Both the AV and the pedestrian are modeled as dynamic-belief-induced quantal cognitive hierarchy (DB-QCH) models, considering human reasoning limitations and bounded rationality in the decision-making process. In addition, a dynamic belief updating mechanism allows the AV to update its understanding of the opponent’s rationality degree in real-time based on observed behaviors and adapt its strategies accordingly. The analysis results indicate that our models effectively simulate vehicle-pedestrian interactions and our proposed AV decision-making approach performs well in safety, efficiency, and smoothness. It closely resembles real-world driving behavior and even achieves more comfortable driving navigation compared to our previous virtual reality experimental data.
\end{abstract}


%
\IEEEpeerreviewmaketitle

\section{Introduction}
\IEEEPARstart{A}{utonomous} vehicle(AV) technology represents a transformative leap in the automotive sector, promising safer, more efficient, and more convenient transportation in the future \cite{r1}. As this technology advances and becomes increasingly integrated into practical applications, AVs will inevitably share the road with other road users, including pedestrians \cite{r2,r3}. However, the complexity and uncertainty inherent in human behavior present significant challenges for AV decision-making and motion planning, especially at unsignalized intersections where pedestrians are involved \cite{r4}. At these intersections, the priority of right-of-way is often ambiguous due to the absence of traffic signals, leading to potential road conflicts that further complicate AV’s decision-making. Moreover, interactions between vehicles and pedestrians are interdependent and coupled. Pedestrians may exhibit various behaviors, like directly crossing the road or hesitating before crossing. In response, the AV must adjust its strategy accordingly. Conversely, AV actions can also influence pedestrian behavior, such as changing their crossing intention or walking speed based on the approaching AV’s movements. Smooth interaction between AVs and pedestrians is essential on urban unsignalized roadways. Hence, this work focuses on AV decision-making in the AV-pedestrian interactions at an unsignalized intersection, where conflicts may arise. \par
Previous studies have investigated vehicle-pedestrian interactions, often relying on statistical methods \cite{r5,r6} or describing interactions as one-time events \cite{r7}. However, pedestrian behavior, characterized by unpredictability and dynamism, presents challenges for such approaches. Their movements can quickly change \cite{r8}, introducing uncertainty into interactions that traditional methods struggle to capture. By considering uncertainty and dynamic interactions, the partially observable markov decision process (POMDP) framework \cite{r9} provides a modeling approach for decision-making challenges that closely mirror real-world conditions. While widely applied to handle complex environments in vehicle-vehicle interactions, its potential in vehicle-pedestrian interactions remains underexplored. \par
Game theory is frequently used to model the interaction between vehicles and pedestrians. However, most studies assume that all players follow the Nash equilibrium \cite{r10,r11}, possessing unlimited computational reasoning ability to compute optimal actions and perfect rationality to execute them, thus maximizing their utility function in decision-making. In reality, individuals often deviate from the Nash equilibrium due to cognitive limitations \cite{r12}, unable to consistently calculate optimal actions or prone to make errors in complex scenarios. Hence, considering human reasoning levels and bounded rationality is essential to develop more accurate models of real-world behaviors. \par
\begin{figure*}[htp]
\centering
\includegraphics[width=0.98\textwidth]{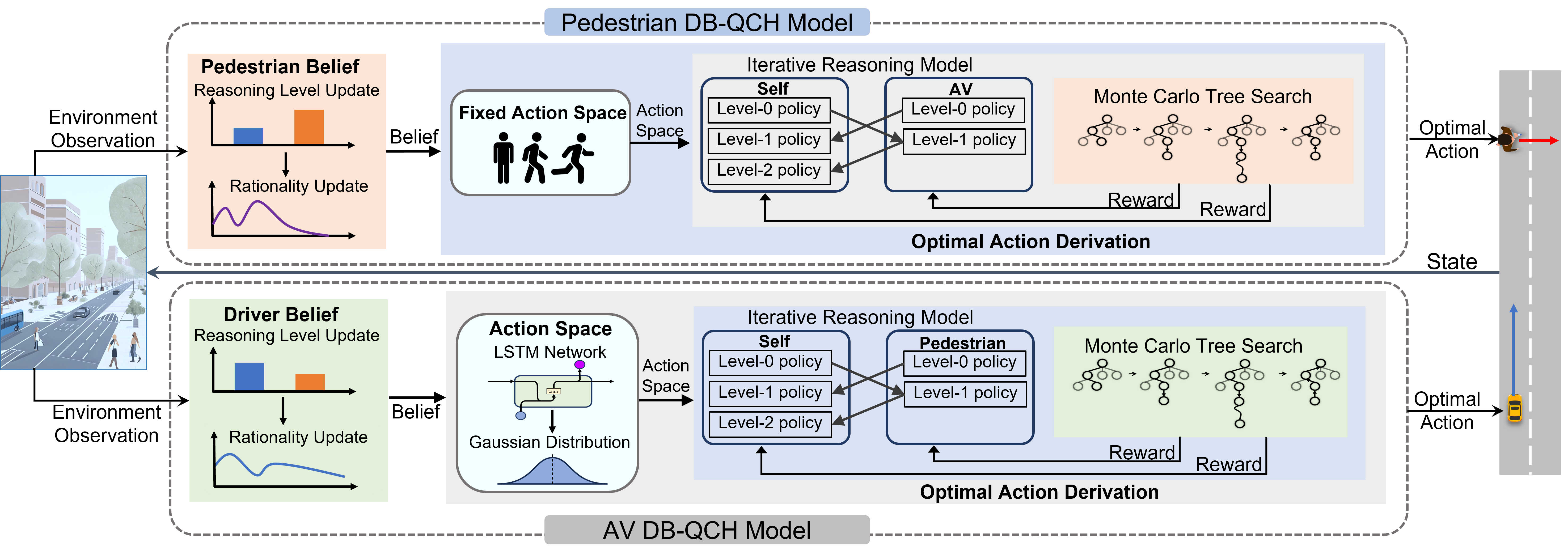}
\captionsetup{belowskip=-10pt} 
\caption{The proposed framework of AV-pedestrian interaction at the unsignalized intersection}
\label{framework}
\end{figure*}

To address these limitations, our study proposes a novel framework that combines POMDP and behavioral game theory to tackle the AV decision-making problem within complex and dynamic environments. Figure \ref{framework} shows our proposed framework for AV-pedestrian interaction at an unsignalized intersection. In this work, we employ the POMDP framework to dynamically model the decision-making process of the AV in an environment with incomplete information and uncertainty. Furthermore, we use a behavioral game theory model to describe AV-pedestrian interaction, both the AV and pedestrian modeled as dynamic-belief-induced quantal cognitive hierarchy (DB-QCH) models. At each time step, the AV model updates its beliefs about its opponent's reasoning level and rationality based on extended Bayesian Estimation. A trained neural network calculates the predicted optimal action, which is then translated to an action space using a Gaussian distribution function. An iterative reasoning model is established to deduce the optimal strategies for both oneself and the opponent at each level, computed via the Monte Carlo Tree Search (MCTS) method. While the pedestrian model is also constructed as a DB-QCH model, their action space remains fixed over time, which differs from the AV model. \par
To enhance humans' understanding of the interaction process and resolve intersection conflicts in a human-involved interactive environment, this work makes the following main contributions: 1). The POMDP framework and behavioral game theory are integrated to address the uncertainty and dynamic interaction between the AV and the pedestrian. 2). To accurately capture the decision-making processes, both the AV and pedestrian are modeled as DB-QCH models. This modeling approach provides a comprehensive understanding of interaction dynamics and facilitates more realistic simulations. 3). A trained neural network based on data from our previous experiments is developed to guide MCTS in exploring the continuous action space of AV, thereby facilitating effective and efficient decision-making. 4). This work introduces variables to quantify human bounded rationality and is the first to propose a dynamic updating mechanism for rational values based on the observed environment, enabling adaptive decision-making by AVs in real-time. These concentrated efforts pave the way for an explainable and trustworthy AV decision-making system, leading to safer and more efficient navigation of AVs in such an interactive environment where humans are involved. \par
The remainder of this paper follows this structure: Section \uppercase\expandafter{\romannumeral2} surveys state of the art in this domain, followed by the problem statement in Section \uppercase\expandafter{\romannumeral3}. Section \uppercase\expandafter{\romannumeral4} outlines the methodology. The experiments and results are thoroughly analyzed and discussed in Section \uppercase\expandafter{\romannumeral5}, and finally, the conclusion is presented in Section \uppercase\expandafter{\romannumeral6}. \par

\section{Related work}
POMDP is a mathematical framework for modeling dynamic systems with imperfect observations, which is an extension of the Markov Decision Process \cite{r13}. In the context of self-driving decision problems, POMDP is commonly employed to capture the incomplete observability and uncertainty in the AV’s surrounding environment. Previous studies have explored various sources of uncertainty in their POMDP models, categorized into offline and online methods based on their solution approaches \cite{r14}. For instance, \cite{r15} regarded pedestrians’ target position as an unobservable variable in the POMDP model to capture the decision-making and planning behavior of autonomous vehicles navigating among many pedestrians. They employed an online planning method to solve this model. Similarly, an AV-pedestrian interaction model was proposed in \cite{r16} to address complex decision-making challenges arising from the uncertain crossing intention of pedestrians in urban environments by leveraging the POMDP framework. In contrast to these approaches, our study treats pedestrian reasoning levels and rationality degrees as unobservable information within the POMDP framework, similar to \cite{r17,r18} focusing on vehicle-vehicle interactions. However, our work innovatively introduces a dynamic mechanism to update rationality levels based on observed behavior and prior knowledge. \par
Game theory \cite{r19} serves as a valuable tool for modeling and analyzing conflicts among individuals, initially used in economics and now extended to vehicle-vehicle or vehicle-pedestrian interactions in the context of AVs. A Zebra Crossing Game was introduced to explore cyclist-vehicle interaction in Norway, demonstrating consistency between real crossing behavior and the solution derived from game theory \cite{r20}. Similarly, the ‘sequential chicken’ model was proposed to simulate space competition between vehicles and pedestrians at an unsigned intersection \cite{r21}. This model was further extended in \cite{r22} by employing empirical data and the Gaussian Process to fit the model’s parameters. A recent study developed a Stackelberg game model based on the belief that players usually make sequential decisions in road conflicts rather than simultaneous responses \cite{r23}. Similar applications of the Stackelberg game model for simulating the interaction between vehicles and pedestrians are observed in \cite{r24} and \cite{r25}. \par
The approaches mentioned earlier are all based on the conventional game theory model with players' complete rationality assumption. However, human behavior does not always conform to the predictions of the Nash equilibrium \cite{r26} in real-world situations due to bounded rationality and cognitive limitation. To relax the assumption of complete rationality, Chen et al. \cite{r28} combined evolutionary game theory with cumulative prospect theory to formulate an interactive decision model at uncontrolled mid-block crosswalks. This method can simulate different behaviors within a pedestrian group but requires numerous parameters for model fitting. \par
In contrast, behavioral game theory provides a more accurate predictor of human behavior in real-world scenarios. It outperforms conventional models in forecasting interaction outcomes \cite{r30}. The researchers argued that Nash equilibrium, normally with complete information may not sufficiently reflect the unpredictable actions of pedestrians at crosswalks. To simulate the joint behavior of pedestrians and vehicles, they proposed a game theoretical framework, namely logit quantal response equilibrium \cite{r31,r5} with incomplete information, replacing Nash equilibrium. Moreover, level-$k$ reasoning softened the perfect rationality assumption of Nash equilibrium by assuming that agents have different levels of reasoning \cite{r33}, applied in diverse vehicle interaction scenarios such as roundabouts \cite{r34}, lane changes \cite{r35}, and intersections \cite{r36}. However, if the opponent's cognitive hierarchy is not at level-$(k-1)$, the level-$k$ model may not perform well in predicting its behavior. Another approach, the ‘cognitive hierarchy' framework, allowed interaction with opponents of varying cognitive levels, not just one level below \cite{r37}.\par
Despite previous efforts using quantal level-$k$ game theory for vehicle-vehicle interactions \cite{r17,r18}, fixed rational levels for humans were a limitation. The quantal cognitive hierarchy model has been demonstrated better performance in predicting human behaviors \cite{r38}. However, its application in the field of autonomous driving remains unexplored. Therefore, our work adopts a DB-QCH model to model the AV and the pedestrian, providing a more accurate description of AV-pedestrian interactions in urban areas. \par

\section{Problem Formulation}
This work focuses on addressing the challenge posed by conflicts arising at the unsignalized intersection, where both the pedestrian and the AV intend to cross simultaneously. Specifically, it aims to develop continuous decision-making strategies for AVs navigating safely and efficiently through such a scenario. As the AV lacks knowledge about the opponent’s intelligence level and rationality in a dynamic and interactive environment, we model the interaction between the AV and the pedestrian using a POMDP framework. The model is defined by the tuple: 
\[ \langle N, S, A, T, O, J, B \rangle \]
\begin{itemize}
    \item \( N = \{0, 1\} \): Represents the two players, where 0 denotes the AV and 1 denotes the pedestrian.
    \item \( S \): A finite set of states, where \( s_t \in S \) signifies the state of the environment at discrete time step \( t \).
    \item \( A = \{A^0, A^1\} \): Defines the action space, with \( A^0 \) representing the AV's actions and \( A^1 \) denoting the pedestrian's actions.
    \item \( T \): The state transition dynamics, expressed as \( s_{t+1} = T(s_t, a^{0}_{t}, a^{1}_{t}) \) for an action pair \( (a^0, a^1) \in A \). This function describes how the environment transitions from one state to another based on the actions of both players.
    \item \( O = \{O^0, O^1\} \): Represents the partially observable state. We assume that each agent's action can be observed, along with certain physical information (e.g., speed, acceleration, distance), while implicit information (e.g., reasoning level, rationality degree) remain unobservable.
    \item \( J = \{J^0, J^1\} \): The utility function for each agent. The utility \( J^{i}_{t} = J^{i}(s_t, a^{0}_{t}, a^{1}_{t}) \), \( i \in N \), depends on both the agent's action and the opponent's action.
    \item \( B = \{B^0, B^1\} \): The belief in the opponent’s intelligence level and rationality, with \( b^{0}_{t} \in B^0 \) and \( b^{1}_{t} \in B^1 \). 
\end{itemize}

For the AV model, the goal is to determine a sequence of optimal actions. The optimization problem can thus be formulated as follows:
\begin{equation}
\begin{aligned}
    & \underset{\pi}{\text{maximize}}
    & & \mathbb{E} \left [\sum_{t=0}^{\infty} \gamma_t J^0_t(s_t, a^0_t, a^1_t, b^0_t) \mid a^0_t\sim\pi  \right] \\
    & \text{subject to}
    & & s_{t+1} = T(s_t, a^{0}_{t}, a^{1}_{t}), \\
    & & & b^0_{t+1} = \rho(b^0_t, o^0_{t}), \\
    & & & a^0_t \in A^0, a^1_t \in A^1, \\
    & & & o^0_{t} \in O^0, b^0_t \in B^0
\end{aligned}
\end{equation}
where $\gamma$ represents the discount factor within the range of $(0, 1]$, while $\rho$ denotes the belief update function.

\section{Methodology}
This section provides a detailed description of the approaches we use to model the interaction between AVs and pedestrians at the unsignalized intersection and solve the above problem.
\subsection{Action Space Generation}
For our AV model, the dynamic decision-making process aims to produce a sequence of expected accelerations. However, in actual scenarios, the AV’s acceleration range exists in a continuous space, posing challenges for methods like MCTS, which typically excel in discrete action spaces. To address this, we employ a pre-trained neural network model to guide MCTS through the continuous action space of AVs. \par
We have opted for the long short-term memory (LSTM) network as our neural network model. LSTM, a subtype of recurrent neural network (RNN), is good at processing and predicting time series data, adeptly capturing temporal dependencies \cite{r40}. Unlike traditional RNNs, LSTM overcomes long-term dependency issues through its gate mechanisms(including input gate, forget gate, and output gate), effectively retaining and leveraging long-term information \cite{r41}. This gives LSTM a significant advantage in handling complex time series data in autonomous driving scenarios.\par
Acceleration prediction in autonomous driving presents a highly temporal problem, as a vehicle's acceleration depends not only on its current state but also on past states and actions. Traditional RNNs often encounter problems like gradient vanishing or exploding when dealing with long-term dependencies \cite{r42}, making it difficult to effectively capture long-term dependency information. LSTM, with its unique gating mechanism, can maintain and transmit key information across lengthy time series, avoiding these shortcomings of traditional RNNs \cite{r43}. \par
Given the temporal dynamics and complexity of acceleration prediction tasks, we choose LSTM as our preferred model for anticipating AV acceleration. Its ability to use historical data enhances prediction accuracy and stability, providing robust support for our decision-making system. \par
Training data for the LSTM model is sourced from our prior vehicle-pedestrian interaction experiments \cite{r44}, conducted using virtual reality(VR) technology. This experiment yielded dynamic interaction data, including the absolute positions of pedestrians and vehicles, vehicle speeds, and driver inputs like steering, throttle, and brakes. For further insights into this VR experiment, please refer to the reference \cite{r44}. Through data processing, we extracted relevant variables such as vehicle speed, acceleration, relative distances, time-to-arrivals, pedestrian speeds, and vehicle yielding status at each time step for every scenario.\par
This data underwent training in the LSTM model, which, post-training, can ingest state information at each time step and output the corresponding anticipated acceleration. This acceleration is treated as the mean of a Gaussian distribution, from which $N$ accelerations are sampled, yielding ${N+1}$ possible accelerations. Subsequently, MCTS is employed to explore these ${N+1}$ actions, leveraging this neural network model's output as an initial guide to enhance MCTS's efficiency in navigating the continuous action space. Through this integration of the neural network and MCTS methods, we can improve the decision-making ability of autonomous vehicles in complex dynamic interactive environments. \par 
In contrast, within the pedestrian model, the action space pertains to the pedestrian's speed. Since pedestrian speed can change rapidly, using a neural network model with Gaussian distribution sampling for action spaces, as used in the AV model, is less effective. Instead, we adopt a discrete action selection space to simplify our model. Starting from 0 m/s, we discretize the speed at 0.1 m/s intervals. This method is simpler than the one used for the AV model and adequately meets our needs. Unlike the AV model, where elements in the action space dynamically change, the pedestrian's action space remains fixed at each time step.\par

\subsection{Dynamic-belief-induced Quantal Cognitive Hierarchy Model}
\subsubsection{Quantal Cognitive Hierarchy model}
The Quantal Cognitive Hierarchy model is a behavioral game theory model used to describe the behavior of bounded rational individuals in games. It integrates the quantal response (QR) model into the traditional cognitive hierarchy (CH) model. \par
In the CH model, agents are characterized by different cognitive levels, each associated with varying degrees of reasoning abilities and consideration of others' behavior. Higher levels indicate greater reasoning capabilities and more consideration of opponents' actions. At each level, agents simulate their opponents' behavior under the assumption that opponents operate at lower levels. Each agent’s cognitive level is denoted by $k$ (where ${k=0, 1, 2…}$). Level-$0$ agents are regarded as non-strategic, generating their strategies independently and without considering opponents’ behavior, often through uniform random selection or simple heuristic methods. Conversely, strategic agents at level-$k$ (where $k > 0$) engage in a more sophisticated decision-making process. They assume their opponents operate at level-$j$, where ${j < k}$, and respond accordingly with optimal strategies. \par
The QR model introduces the concept of bounded rationality, where agents do not always choose the optimal strategy but select strategies with certain probabilities based on expected payoffs when making decisions. In this model, bounded rationality is represented by the parameter $\lambda$ (where $\lambda$ $\in$ [0, $\infty$)), which measures the degree of rationality. A higher $\lambda$ indicates more rational behavior, while a lower value reflects greater randomness in decision-making. The probability $P$($a_i$) that an agent $i$ chooses a particular strategy $a_i$ given the opponent’s action is described by the quantal response function:
\begin{equation}
P(a_i) = \frac{e^{\lambda Q(a_i, a_{-i})}}{\sum_{a'_i \in A} e^{\lambda Q(a'_i, a_{-i})}}
\label{qrf}
\end{equation}
where $Q_i(a_i, a_{-i})$ is the expected payoff for agent $i$ when choosing strategy $a$. \par
Equation \ref{qrf} shows that the probability of selecting a strategy increases with its expected payoff, meaning individuals are more likely to select a strategy with a higher expected payoff but may also opt for those with lower returns. As $\lambda$ approaches infinity, the model approximates perfect rationality, where the highest payoff strategy is always chosen. Conversely, when $\lambda$ is close to zero, the choice of strategy becomes completely random.\par 
By combining ideas from the CH model and QR model, the QCH model offers insights into how individuals probabilistically select strategies at different cognitive levels, thereby enhancing our understanding of bounded rational behavior. In our study, we adopt the QCH model to represent the decision-making process for both the AV and the pedestrian. This model captures the varying levels of intelligence $k$ and rationality $\lambda$ of each opponent, which are unobservable to each other. Here, we detailed the decision-making process specifically using the AV QCH model as an example. \par
At each level-$k$, the AV evaluates its potential actions by calculating the expected payoff $Q$ for each action given the current state $s_t$. This evaluation also considers the pedestrian's policy from the preceding level-($k-1$). The AV then makes a quantal best response to the pedestrian's level-($k-1$) policy. In addition, the AV model reasons the potential actions available to the pedestrian at this level, preparing for subsequent policy computations. The policy at each level for the AV and its opponent is developed through a sequential and iterative process, starting from level-$0$ to higher levels. In our study, we assume that the level-$0$ agent lacks understanding of pedestrian intentions or higher-level policies, and instead treats pedestrians as stationary obstacles to compute its actions. In contrast, the level-$k$ (where $k > 1$) agent regards its opponents as level-$(k-1)$ agents. Specifically, the quantal response function is used to compute the policy:
\begin{equation}
\pi^{i, k, \lambda^i}(a^i_j) = \frac{e^{\lambda^i Q^{k}(s_t, a^i_j, \pi^{-i, k-1, \lambda^{-i}})}}{\sum_{a' \in A^i} e^{\lambda^i Q^{k}(s_t, a', \pi^{-i, k-1, \lambda^{-i}})}}
\end{equation}
After computing strategies for all levels, we can derive the optimal strategy using initial beliefs. Finally, we can determine the optimal action for the AV, selecting the action associated with the highest mixed strategy value. The algorithm for this process is shown in Algorithm \ref{alg1}. \par

\begin{algorithm}[htbp]
\caption{QCH Model Iterative Reasoning to Compute Optimal Action}\label{alg1}
\KwIn{
    $N$: Player set, $A$: Possible action set, $s_t$: Current state, $\pi^{i,0}$: The level-0 policy for agent $i$, $K$: Maximum cognitive level, $b_k$: Belief about the opponent’s level, $\lambda$: Rationality degree}
\KwOut{Optimal action}
Initialize $agent\_policy \gets []$\;
Initialize $mix\_policy \gets []$\;
Append $\pi^{i,0}$ to $agent\_policy$\;
\For{$k = 1$ \KwTo $K$}{
    \For{each player $i \in N$}{
        \For{each action $a^i_j \in A^i$}{
            Compute the payoff $Q^{i,k}(s_t, a^i_j, \pi^{-i, k-1, \lambda^{-i}})$\;
        }
        Compute the policy $\pi^{i, k, \lambda^i}(a^i_j) = \frac{e^{\lambda^i Q^{i,k}(s_t, a^i_j, \pi^{-i, k-1, \lambda^{-i}})}}{\sum_{a' \in A^i} e^{\lambda^i Q^{i,k}(s_t, a', \pi^{-i, k-1, \lambda^{-i}})}}$\;
        Append $\pi^{i, k, \lambda^i}$ to $agent\_policy$\;
    }
}
\For{$k = 1$ \KwTo $K$}{
    $mix\_policy \gets mix\_policy + b_k[k-1] \cdot agent\_policy[k]$\;
}
$optimal\_index \gets \arg\max(\text{$mix\_policy$})$\;
$optimal\_action \gets A[optimal\_index]$\;
\Return $optimal\_action$\;
\end{algorithm}
Introducing the QCH model provides the advantage of simultaneously accounting for the agent's limited rationality and reasoning level, thus making the model more realistic. Through this approach, we can more accurately simulate the decision-making behaviors of real-world agents and gain insights into the interactions between AVs and pedestrians. \par

\subsubsection{Dynamic Belief Update}
For AVs, the opponent's reasoning level-$k$ and rationality degree $\lambda$ are not directly observable. Pedestrian behavior is dynamic and may constantly change. If AVs always use fixed values of pedestrian’s cognitive states for best response calculation during interactions with pedestrians, they will be unable to effectively identify and adapt to changes in pedestrian behavior. This would make AVs appear less intelligent and human-like, as humans can quickly recognize and respond to sudden behaviors. The Bayesian approach allows AVs to continuously learn and update their beliefs about pedestrians' reasoning level-$k$ and rationality $\lambda$ during interactions.\par
At time step $t=0$, the agent $i$ establishes an initial belief $b_{k,0}$  about the pedestrian's reasoning level, according to the initial environmental state and our prior experimental data on human-vehicle interactions \cite{r44}. Throughout the game reasoning process, its QCH model iteratively predicts the expected utility of the opponent's potential actions across each reasoning level $k$ for the next state $s_{t+1}$, alongside computing the associated probability $P(s_{t+1}, a^{-i}_{t+1}|k)$, where $s_{t+1} \in S, a^{-i}_{t+1} \in A^{-i}, k \in \mathbb{N}$ for each action. Upon observing the opponent's latest action $a^{-i}_{t+1}$ at time step $t+1$, the agent model identifies the probability value of the action that is closest to the actual observed action  at each reasoning level $k$.  Subsequently, followed by normalizing the probabilities to ensure coherence, it updates its belief $b_{k,t+1}$ concerning its opponent's reasoning level across all levels using the Bayesian equation:
\begin{equation}
P(k|s_{t+1}, a^{-i}_{t+1}) = \frac{P(s_{t+1}, a^{-i}_{t+1}|k) b_{k,t}(k)}{\sum_{k' \in \Theta} P(s_{t+1}, a^{-i}_{t+1}|k') b_{k,t}(k')}
\end{equation}
where $P(k|s_{t+1}, a^{-i}_{t+1})$ $\in$ $b_{k,t+1}$, and $\Theta$ represents all possible values for the reasoning level. \par
Our model differs from others by dynamically updating the belief about the opponent’s rationality degree $\lambda$, rather than relying on constant values. The Bayesian approach is employed to capture changes in the agent $i$’s understanding of its opponent in this parameter. Initially, agent $i$ has prior knowledge regarding the distribution of $\lambda$, denoted as $f_t(\lambda)$, indicating its estimation of the opponent $j$’s rationality at the current time step $t$. Since $\lambda$ $\in$ [0, $\infty$), it is treated as a continuous variable. Therefore, we use a Bayesian updating method suitable for continuous variables \cite{r47}: 
\begin{equation}
f_{t+1}(\lambda|a^j_t) = \frac{P(a^j_t|\lambda) f_t(\lambda)}{\int_0^\infty P(a^j_t|\lambda')| f_t(\lambda') d\lambda'}
\end{equation}

Considering the varying reasoning level of agent $j$, the actions it takes will correspond to its specific reasoning level. Consequently, we can extend the Bayesian formula accordingly: \par
\begin{equation}
f_{t+1}(\lambda|a^j_t,k) = \frac{P(a^j_t|k,\lambda) f_t(\lambda)}{\int_0^\infty P(a^j_t|k, \lambda')| f_t(\lambda') d\lambda'}
\label{eqft1}
\end{equation}

Since it involves integrating the function value over an infinite interval, it is challenging to compute Equation \ref{eqft1} after multiple iterations. The conjugate prior distribution proves highly effective in addressing this issue \cite{r48}. In Bayesian methodology, if the posterior distribution belongs to the same family as the prior distribution, the prior distribution is referred to as a conjugate prior distribution. The benefit of using a conjugate prior lies in its ability to simplify the Bayesian update process, requiring only the adjustment of parameters within the conjugate prior distribution to complete the Bayesian inference update. Our work considers the following family of distributions \cite{r49}:
\begin{equation}
\resizebox{\linewidth}{!}{$ 
f(\lambda;Q,n_0,n_1,\ldots,n_K) = \frac{e^{\lambda Q}/\prod_{k=0}^{K}(\sum_{l=1}^{m}e^{\lambda Q_{a_{l, k}}})^{n_k}}{\int_0^\infty e^{\lambda' Q}/\prod_{k=0}^{K}(\sum_{l=1}^{m}e^{\lambda' Q_{a_{l, k}}})^{n_k} d\lambda'}
$}
\label{eqf}
\end{equation}
where $n_k \in \mathbb{N}, \forall k = 0, 1, 2, \ldots, K$, representing the number of occurrence agent $j$'s reasoning level corresponds to $k$. $f(\lambda;Q,n_0,n_1,\ldots,n_K)$ is a probability density function as $\int_0^\infty f(\lambda;Q,n_0,n_1,\ldots,n_K)d\lambda = 1$. \par

At each time step $t$, through game reasoning, agent $i$ calculates the expected utility $Q_{a_{j,k}}$ for each action that their opponent may take at each level and predicts the reasoning level $k$ of agent $j$. By using the prior distribution $f(\lambda;Q,n_0,n_1,\ldots,n_K)$ and the observed action $a_j$ taken by its opponent, agent $i$ can then update the distribution of $\lambda$, $f(\lambda;Q+Q_{a_{j,k}},n_0,n_1,\ldots,n_k+1,\ldots, n_K)$, and use it as the prior distribution at the next time step $t+1$. It is known that when employing Equation \ref{eqf} as the prior distribution, updating the parameters $Q$ and $n_k$ to obtain the posterior distribution $f(\lambda;Q+Q_{a_{j,k}},n_0,n_1,\ldots,n_k+1,\ldots, n_K)$ can be directly achieved without the need for intricate integral calculations. \par 
\textbf{Theorem 1:} Given the prior distribution $f_t(\lambda;Q,n_0,n_1,\ldots,n_K)$, upon observing the action $a_{j}$ taken by the opponent at time step $t+1$, agent $i$ can update the belief as $f_{t+1}(\lambda;Q+Q_{a_{j,k}},n_0,n_1,\ldots,n_k+1,\ldots, n_K)$.\par
\textbf{Proof:}
\begin{align*}
f(\lambda|a_{j,k}) &=  \frac{P(a_{j,k}|\lambda) f(\lambda;Q,n_0,n_1,\ldots,n_K)}{\int_0^\infty P(a_{j,k}|\lambda') f(\lambda';Q,n_0,n_1,\ldots,n_K) d\lambda'} \\
&= \frac{\frac{e^{\lambda Q_{a_{j,k}}}}{\sum_{l=1}^{m}e^{\lambda Q_{a_{j,k}}}} \cdot \frac{e^{\lambda Q} \cdot g(Q,n)}{\prod_{k=0}^{K}(\sum_{l=1}^{m}e^{\lambda Q_{a_{l, k}}})^{n_k}}}{\int_0^\infty \frac{e^{\lambda' Q_{a_{j,k}}}}{\sum_{l=1}^{m}e^{\lambda' Q_{a_{j,k}}}} \cdot \frac{e^{\lambda' Q} \cdot g(Q,n)}{\prod_{k=0}^{K}(\sum_{l=1}^{m}e^{\lambda' Q_{a_{l, k}}})^{n_k}} d\lambda'} \\
&= \frac{\frac{e^{\lambda(Q_{a_{j,k}}+Q)} \cdot g(Q,n)}{\prod\limits_{\substack{i=0 \\ i \neq k}}^{K} (\sum_{l=1}^{m}e^{\lambda Q_{a_{l,k}}})^{n_k} \cdot (\sum_{l=1}^{m} e^{\lambda Q_{a_{l,k}}})^{n_{k}+1}}}   {\int_0^\infty \frac{e^{\lambda'(Q_{a_{j,k}}+Q)} \cdot g(Q,n)}{\prod\limits_{\substack{i=0 \\ i \neq k}}^{K} (\sum_{l=1}^{m}e^{\lambda' Q_{a_{l,k}}})^{n_k} \cdot (\sum_{l=1}^{m} e^{\lambda' Q_{a_{l,k}}})^{n_{k}+1}} d\lambda'} \\
&= f(\lambda;Q+Q_{a_{j,k}},n_0,n_1,\ldots,n_k+1,\ldots, n_K)
\end{align*}
where we denote $\frac{1}{\int_0^\infty e^{\lambda' Q}/\prod_{k=0}^{K}(\sum_{l=1}^{m}e^{\lambda' Q_{a_{l, k}}})^{n_k} d\lambda'}$ as $g(Q,n)$ for simplicity.\par
\bigskip
When the distribution of the continuous variable $\lambda$ at the next time step $s_{t+1}$ is obtained, the expectation of rationality degree can be calculated using the following equation: 
\begin{equation}
 \mathbb{E}(\lambda) = \int_0^\infty \lambda f(\lambda;Q+Q_{a_{j,k}},n_0,n_1,\ldots,n_k+1,\ldots, n_K) d\lambda
\end{equation}

The above describes how to update the knowledge of the opponent's reasoning level $k$ and rationality degree $\lambda$ at each time step. Algorithm \ref{alg2} provides a clearer understanding of the process. \par
Through dynamic belief updates, AVs can more accurately predict pedestrian behavior and adjust their strategies based on the latest beliefs. This dynamic update mechanism enables AVs to better adapt to changing environments, thereby behaving more intelligently and human-like. \par
\begin{algorithm}[htbp]
\caption{Belief Update}\label{alg2}
\KwIn{
    $s_t$: Current state, $a_j$: Observed opponent's action, $Q_{a_{j,k}}$: Opponent's action $a_j$'s expected utility for each level $k$, $b_{k,t}$: Prior belief about reasoning level, $P(s_{t+1},a_j|k)$: Probability for $a_j$ at each level $k$, $f(\lambda;Q,n_0,n_1,\ldots,n_K)$: Prior distribution about rationality}
\KwOut{Updated belief $b_{k, t+1}$, $b_{\lambda, t+1}$}
\For{$k = 0$ \KwTo $K-1$}{
    $P(k|s_{t+1}, a_{j,t+1}) = \frac{P(s_{t+1}, a_{j,t+1}|k) b_{k,t}(k)}{\sum_{k' \in \Theta} P(s_{t+1}, a_{j,t+1}|k') b_{k,t}(k')}$    \\
    $b_{k, t+1}(k) \gets P(k|s_{t+1}, a_{j,t+1})$\;
    }
$k \gets \arg\max(b_{k,t+1})$\;
$\mathbb{E}(\lambda) \gets \int_0^\infty \lambda f(\lambda;Q+Q_{a_{j,k}},n_0,\ldots,n_k+1,\ldots, n_K) d\lambda$\;
$b_{\lambda, t+1} \gets \mathbb{E}(\lambda)$\;
\Return $b_{k, t+1}$, $b_{\lambda, t+1}$\;
\end{algorithm}
\vspace{-5pt}

\subsection{MCTS}
Monte Carlo Tree Search is a heuristic search algorithm to predicts future outcomes and optimizes decision-making by simulations. It includes four main steps: selection, expansion, simulation, and backpropagation \cite{r45}. In our study, MCTS is used to compute the anticipated payoff of each possible action for both the AV and pedestrian models at each level for each moment.\par
Specifically, the decision tree is initialized at every time step from the current state $s_t$. In the selection stage, we use the Upper Confidence Bound applied to Trees (UCT) formula to calculate the UCT value for each potential action in the action space and select the one with the highest UCT value for the next expansion. The UCT equation is shown below \cite{r46}:
\begin{equation}
\text{UCT}(s,a) = \bar{Q}_a + C \sqrt{\frac{\ln N_s}{N_a}}
\end{equation}
where \(\bar{Q}_a\) is the average utility of action $a$ at state $s_t$, $N_s$ is the total number of visits to the state $s$, $N_a$ is the number of times action $a$ was chosen, and $C$ is a constant that balances exploration and exploitation. \par
During the expansion step, new decision nodes are generated, corresponding to different actions that the AV and the pedestrian may take in the current state. The simulation step starts from the newly expanded node, where the actions of the agent and its opponent are randomly simulated in turn until reaching the terminal state or the maximum search depth. In this stage, a random strategy is used to select actions and simulate the opponent’s strategies, estimating the potential value of the node. Notably, when calculating the optimal strategy for the agent at level-$k$, its opponent follows the policy of level-$k-1$; for level-$0$ agents, their opponents are regarded as static obstacles. Upon simulation completion, the rewards obtained are backpropagated to the root node, updating the node statistics information. After multiple iterations, the average cumulative utility for each action will be utilized to calculate the quantal response policy. \par
Through this process, MCTS evaluates the potential effects of different actions through extensive simulations without relying on specific domain knowledge, enabling AV to make efficient and safe decisions in complex and uncertain environments. \par

\section{Experiments and results}
\subsection{Experiment setup}
We conducted a series of simulation experiments for verification to evaluate the effectiveness of the AV model and decision-making algorithm we developed. We built a simulation scenario where the AV interacts with a pedestrian crossing the road. The AV is 5 m long and 2 m wide, driving on a single-lane road that is 3.65 m wide. The pedestrian's goal is to cross the road to reach the opposite side. \par
We previously conducted experiments on real human drivers and pedestrians interacting in a VR environment \cite{r44}. During these experiments, the driver continuously drove along the road at a random speed and interacted with the pedestrian crossing the road. Since VR technology providing reliable and essential data has been widely applied in the interactive research and testing of AVs \cite{r50, r51}, we are expected to validate and evaluate our model by comparing our simulation data with the VR experimental data. We randomly selected 100 scenarios from these experimental data for this simulation experiment. The initial conditions of each scenario were input into our model for simulation, and each scenario was simulated 100 times to ensure the reliability and statistical significance of the results. \par
The initial data included the initial positions of the vehicle and the pedestrian, the vehicle’s initial velocity and acceleration, and the pedestrian’s velocity. In our models, we assume that the AV and pedestrian follow a straight line with only longitudinal movements. The acceleration range of the AV is set to [-5, 5] m/s$^{2}$, and the speed range of the pedestrian is [0, 0.1, 0.2, 0.3, 0.4, 0.5, 0.6, 0.7, 0.8, 0.9, 1.0, 1.1, 1.2] m/s. We assume that the maximum reasoning ability of the agent is $k$\_{max}= 2, with an initial rationality degree value of $E(\lambda)=$ 10. Each time step is 0.8 s. The level-0 policy is non-strategic, which believes that the agent calculates its optimal strategy based on the assumption that its opponents are static obstacles. \par

\subsection{Results}
This section will verify and evaluate our proposed model's performance through qualitative and quantitative analysis. \par
\subsubsection{Qualitative analysis}
Three simulation examples will be given to clearly and intuitively demonstrate how our AV model operates under different scenarios.\par
\begin{figure}[htp]
\centering
\includegraphics[width=0.49\textwidth]{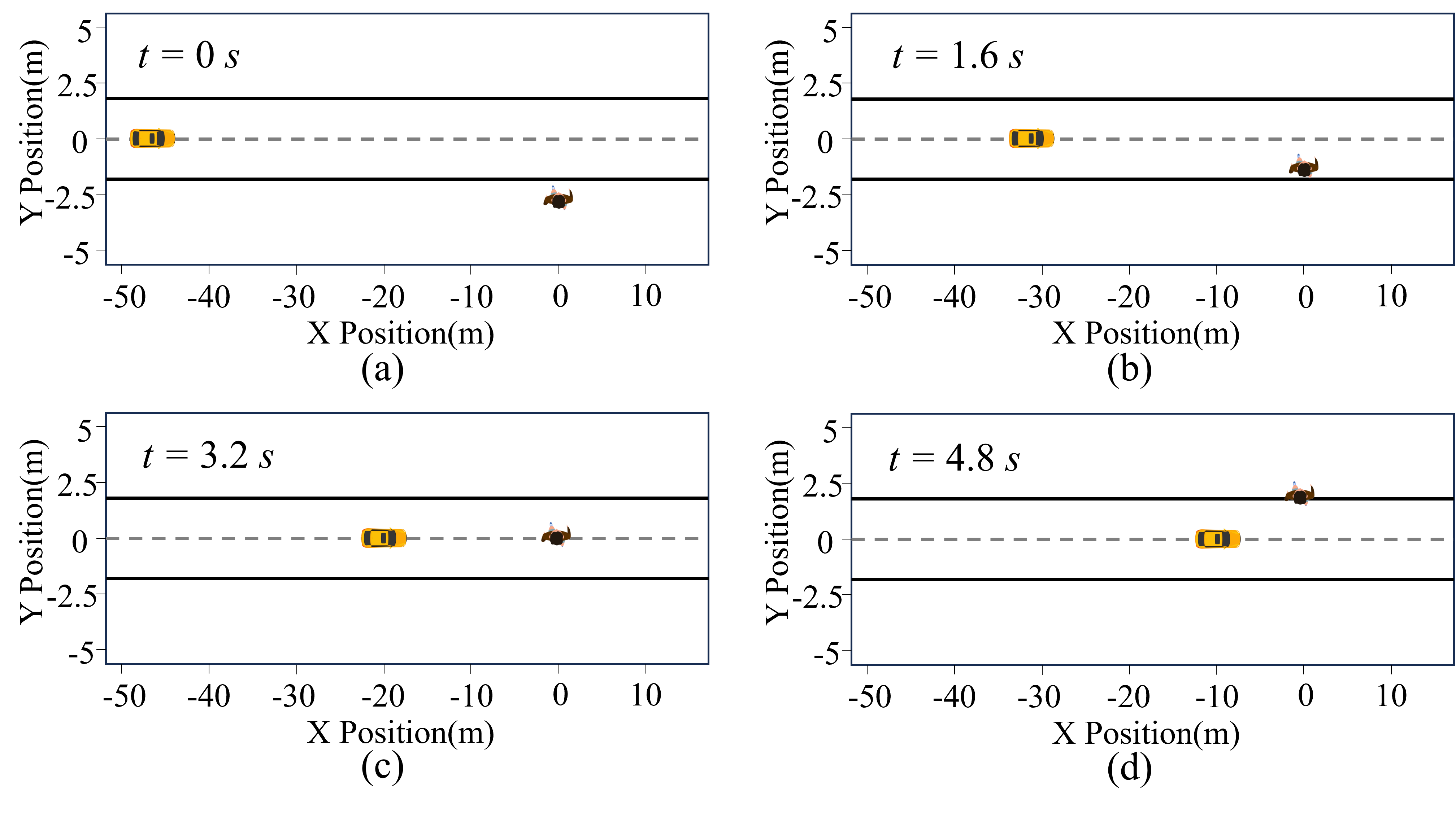}
\captionsetup{belowskip=-10pt}
\caption{Simulation of interaction process in case 1. (a) State: $t$ = 0 s, $v_{\text{ped}}$ = 0.03 m/s, $v_{\text{AV}}$ = 9.348 m/s, $a_{\text{AV}}$ = -0.11 m/s$^2$. (b) State: $t$ = 1.6 s, $v_{\text{ped}}$ = 1.2 m/s, $v_{\text{AV}}$ = 8.31 m/s, $a_{\text{AV}}$ = -1.397 m/s$^2$. (c) State: $t$ = 3.2 s, $v_{\text{ped}}$ = 1.2 m/s, $v_{\text{AV}}$ = 6.72 m/s, $a_{\text{AV}}$ = -0.111 m/s$^2$. (d) State: $t$ = 4.8 s, $v_{\text{ped}}$ = 0.7 m/s, $v_{\text{AV}}$ = 6.57 m/s, $a_{\text{AV}}$ = -0.283 m/s$^2$.}
\label{d46}
\end{figure}
\begin{figure}[htp]
\centering
\includegraphics[width=0.49\textwidth]{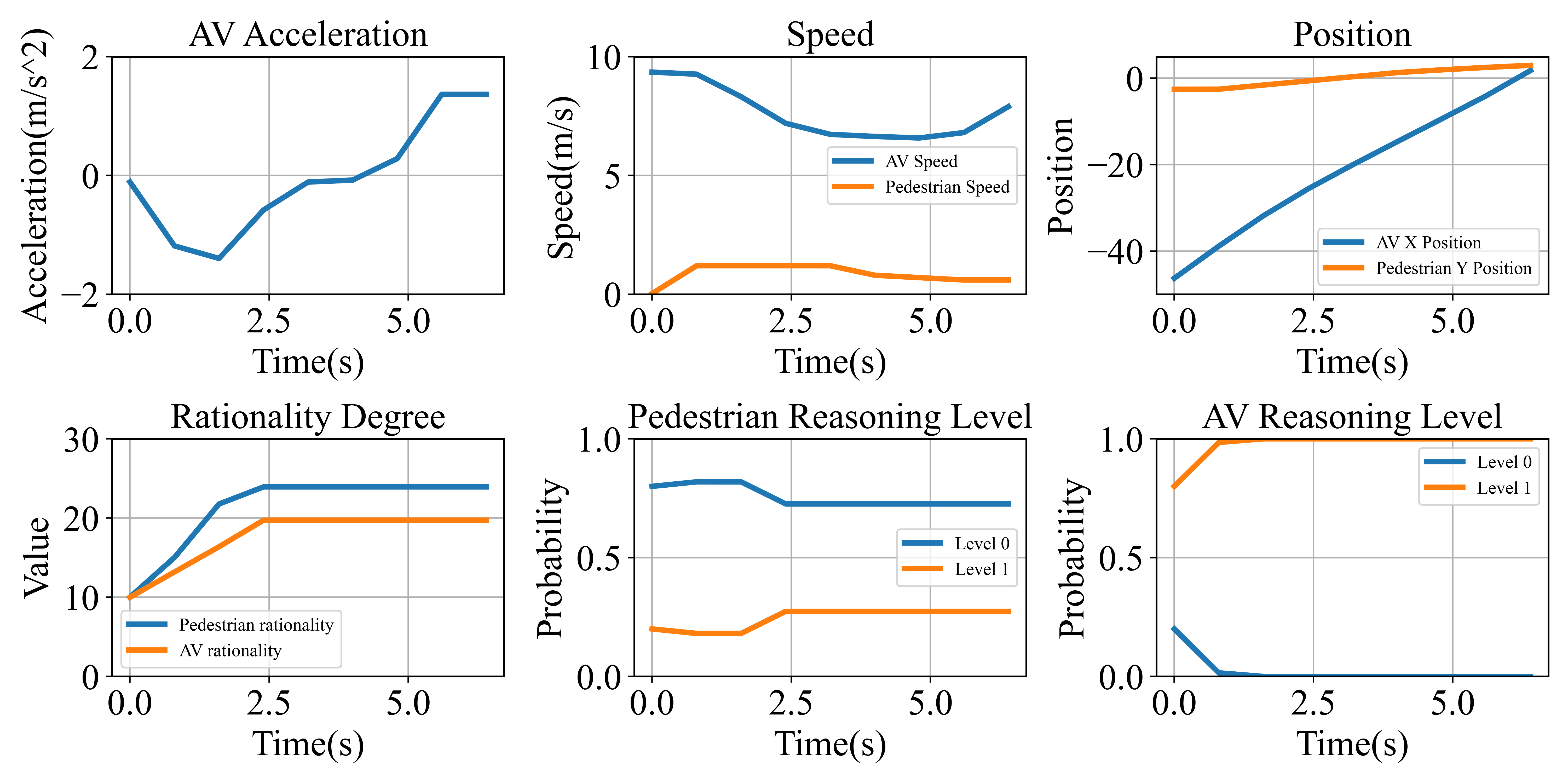}
\captionsetup{belowskip=0pt}
\caption{State evolution in the simulation of case 1.}
\label{d46figure}
\end{figure}

\textbf{\textit{Case 1}}: This case illustrates an AV yielding to a pedestrian, randomly selected from a set of 100 scenarios. Initially, the AV is 46.309 m away from the pedestrian, traveling at a speed of 9.348 m/s with an acceleration of -0.11 m/s$^2$ towards the pedestrian. The pedestrian is located 2.564 m laterally from the AV on the road. Figure \ref{d46} shows the interaction process, combining visual snapshots of their positions with detailed speed data. Information about the state evolution of the pedestrian and vehicle is provided in Figure \ref{d46figure}.
Under these initial conditions, the AV calculates a higher probability that the pedestrian is a level-$0$ agent, indicating that the pedestrian is more likely to cross the road. Therefore, the AV decides to decelerate to avoid collision. When the AV observes the pedestrian stepping into the lane at a speed of 1.2 m/s, it executes a more pronounced deceleration. At the same time, the AV updates its assessment of the pedestrian's rationality based on their actions, concluding that the pedestrian remains in a rational state. \par
As the pedestrian continues to cross and approach its destination, the AV gradually reduces its deceleration and eventually transitions back to acceleration. Notably, the AV does not come to a complete stop but maintains a reduced speed while the pedestrian crosses. This entire process demonstrates the AV’s ability to yield to pedestrians by making real-time decisions, ensuring both safety and efficiency. \par
\begin{figure}[htp]
\centering
\includegraphics[width=0.49\textwidth]{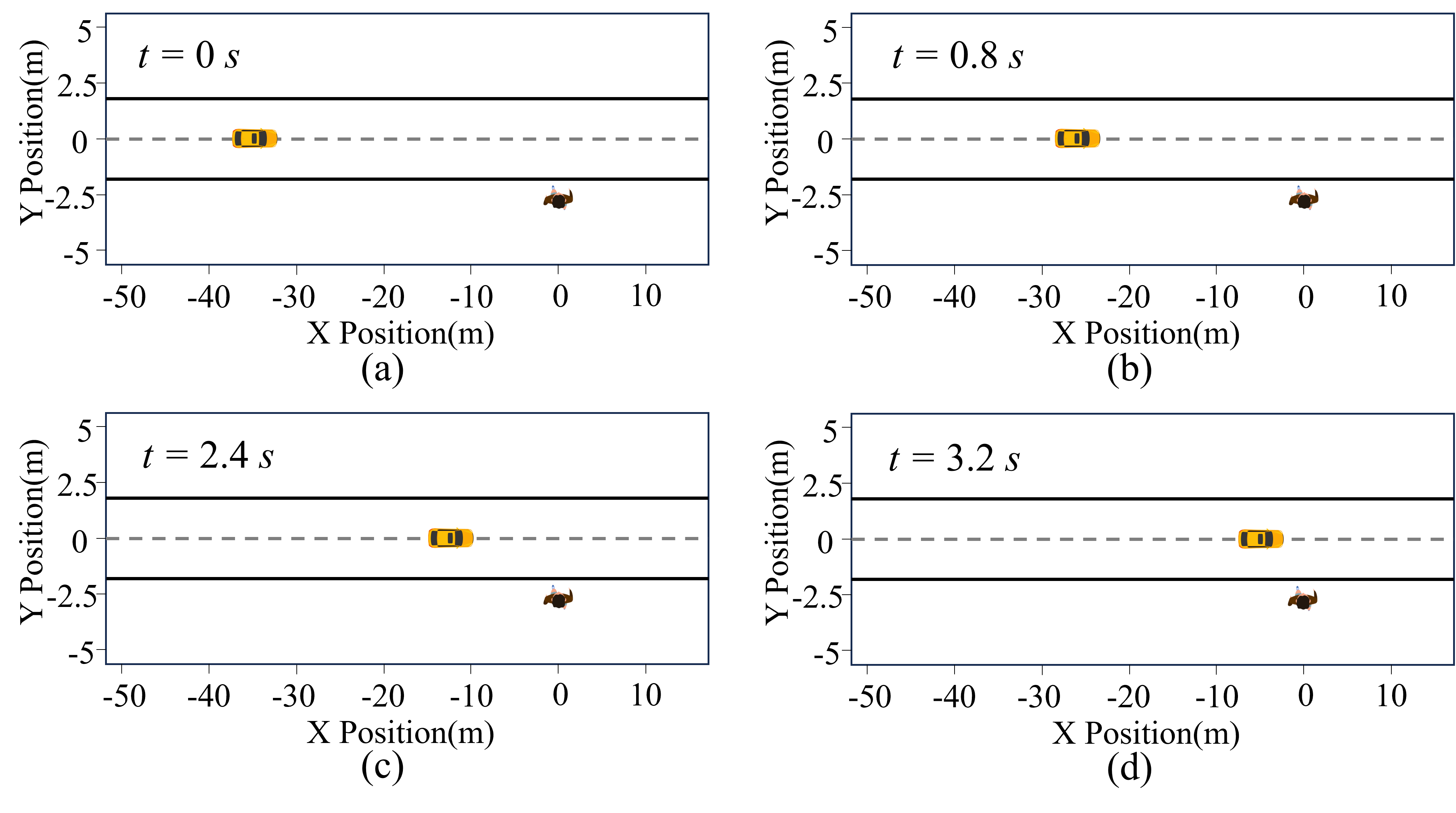}
\captionsetup{belowskip=-10pt}
\caption{Simulation of interaction process in case 2. (a) State: $t$ = 0 s, $v_{\text{ped}}$ = 0.03 m/s, $v_{\text{AV}}$ = 9.348 m/s, $a_{\text{AV}}$ = -0.11 m/s$^2$. (b) State: $t$ = 0.8 s, $v_{\text{ped}}$ = 0 m/s, $v_{\text{AV}}$ = 9.26 m/s, $a_{\text{AV}}$ = -0.33 m/s$^2$. (c) State: $t$ = 2.4 s, $v_{\text{ped}}$ = 0 m/s, $v_{\text{AV}}$ = 8.83 m/s, $a_{\text{AV}}$ = -0.216 m/s$^2$. (d) State: $t$ = 3.2 s, $v_{\text{ped}}$ = 0.1 m/s, $v_{\text{AV}}$ = 8.62 m/s, $a_{\text{AV}}$ = -0.175 m/s$^2$.}
\label{d34}
\end{figure}
\begin{figure}[htp]
\centering
\includegraphics[width=0.49\textwidth]{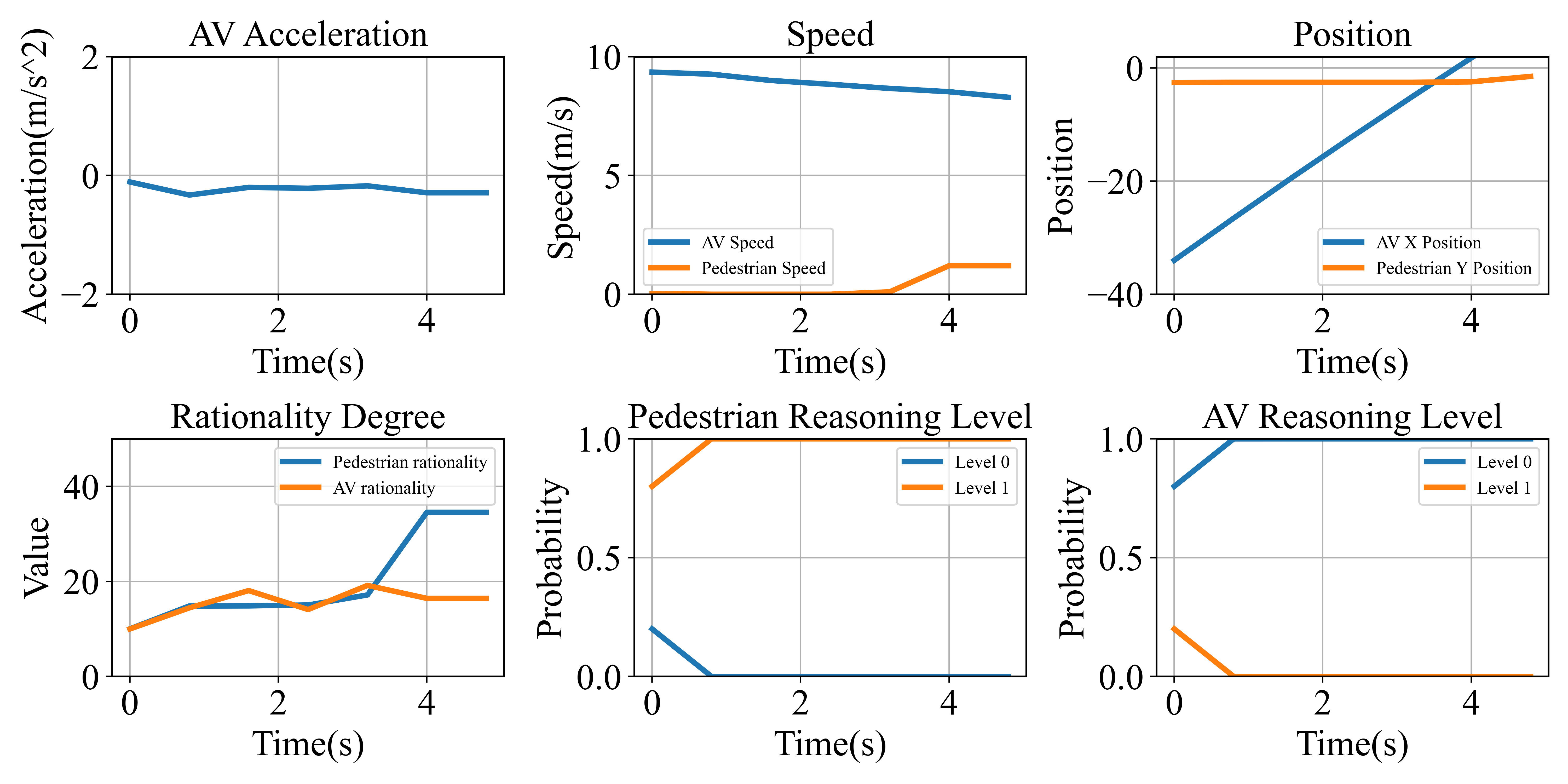}
\captionsetup{belowskip=-5pt}
\caption{State evolution in the simulation of case 2.}
\label{d34figure}
\end{figure}

\textbf{\textit{Case 2}}: This case illustrates a scenario where a pedestrian yields to an AV. Compared to case 1, the initial longitudinal distance between the AV and the pedestrian is updated to 34 m with other conditions remaining the same. The interaction process and detailed evolution states are shown in Figures \ref{d34} and \ref{d34figure}. \par
At the first time step, the AV is 34 m away, moving at a speed of 9.348 m/s, and an acceleration of -0.11 m/s$^2$. The pedestrian is positioned on the road at a lateral distance of 2.564 m from the AV. Unlike case 1, the AV determines that the pedestrian poses a lower probability of crossing and designates the pedestrian as a level-$1$ agent, signifying that the pedestrian will not recklessly enter the lane and yield to AV. \par
As a result, the AV opts to maintain a slight deceleration in case the pedestrian suddenly changes his mind and steps into the lane. However, after observing that the pedestrian does not appear in the lane, the AV confirms that the pedestrian will indeed not cross and perceives its behavior as rational. At this point, the AV sustains a slight deceleration to ensure the pedestrian's safety and maintain smooth operation without significant speed adjustments. \par
\textbf{\textit{Case 3}}: This case primarily aims to validate the AV's ability to identify irrational pedestrian behavior. In the initial conditions of Case 2, the AV deemed the pedestrian's decision to yield as rational, and the pedestrian model indeed waited for the AV to pass before crossing the road. In this case, we maintain the same initial conditions as in Case 2, with the only difference being the replacement of the pedestrian model with custom-design action for the pedestrian. Specifically, we program the pedestrian to cross the road under these conditions at a speed of 1.2 m/s.\par
\begin{figure}[H]
\centering
\includegraphics[width=0.49\textwidth]{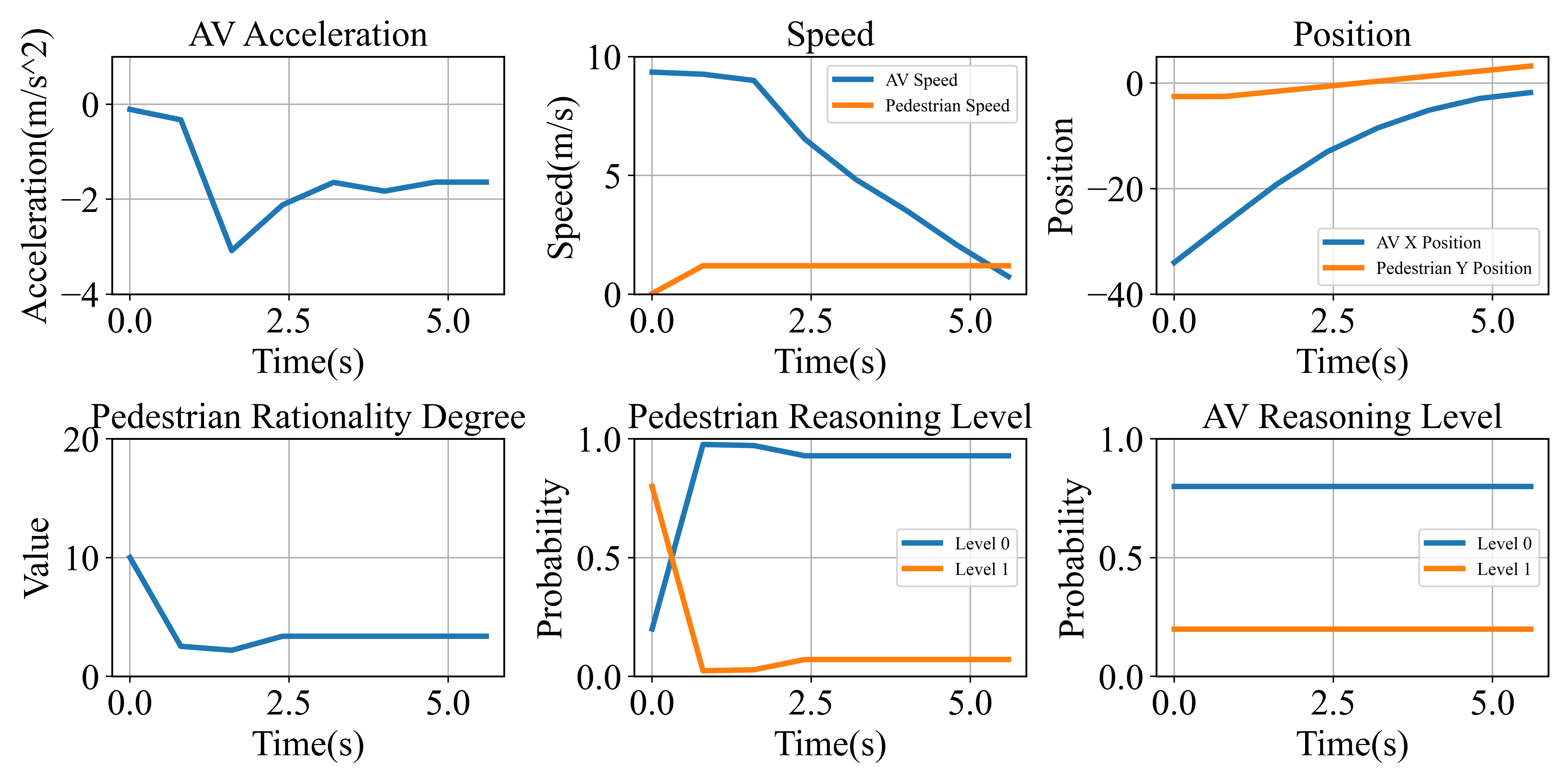}
\captionsetup{belowskip=0pt}
\caption{State evolution in the simulation of case 3.}
\label{d34figurefixedcase}
\end{figure}

As depicted in Figure \ref{d34figurefixedcase}, the AV, similar to Case 2, initially assumes a low likelihood of pedestrians crossing the road and thus only slows down slightly. However, when the pedestrian begins moving, the AV updates its estimation of the pedestrian being a level-$0$ agent based on their actions. At the same time, a marked drop in the pedestrian’s rationality value indicates that the AV deems it irrational for the pedestrian to cross the road under the current circumstance. According to these judgments, the AV decelerates more sharply than in Case 2, resulting in a rapid decrease in vehicle speed. This indicates that the proposed AV model can effectively update its understanding of pedestrian rationality based on real-time behaviors and appropriately adjust its acceleration to prevent potential collisions.\par
\vspace{-5pt}
\begin{figure}[H]
\centering
\includegraphics[width=0.49\textwidth]{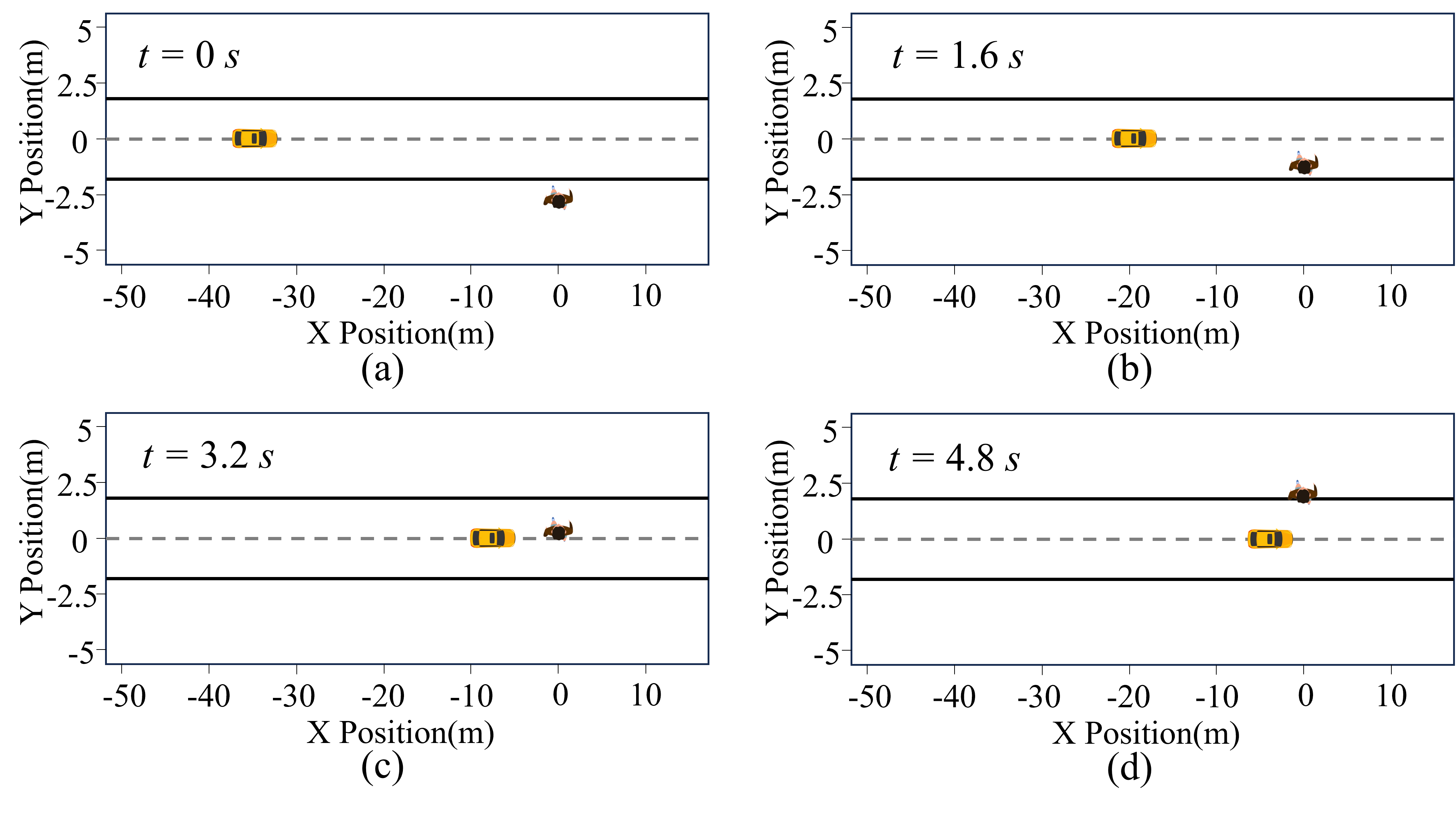}
\captionsetup{belowskip=0pt}
\caption{Simulation of interaction process in case 3. (a) State: $t$ = 0 s, $v_{\text{ped}}$ = 0.03 m/s, $v_{\text{AV}}$ = 9.348 m/s, $a_{\text{AV}}$ = -0.11 m/s$^2$. (b) State: $t$ = 1.6 s, $v_{\text{ped}}$ = 1.2 m/s, $v_{\text{AV}}$ = 8.99 m/s, $a_{\text{AV}}$ = -3.081 m/s$^2$. (c) State: $t$ = 3.2 s, $v_{\text{ped}}$ = 1.2 m/s, $v_{\text{AV}}$ = 4.83 m/s, $a_{\text{AV}}$ = -1.646 m/s$^2$. (d) State: $t$ = 4.8 s, $v_{\text{ped}}$ = 1.2 m/s, $v_{\text{AV}}$ = 2.05 m/s, $a_{\text{AV}}$ = -1.641 m/s$^2$.}
\label{d34bad}
\end{figure}
\vspace{-18pt}
\vspace{-8pt}
In conclusion, the AV's ability to dynamically adjust its perception of pedestrian behavior and response with suitable deceleration shows a good performance of the AV model in various pedestrian interaction scenarios.\par

\subsubsection{Quantitative analysis}
Following the analysis method in the work \cite{r27}, we will conduct the quantitative evaluation from three aspects: safety, efficiency, and smoothness. We input 100 scenarios into our model for simulation, with each scenario being simulated 100 times. Results in Table \ref{tab1} show the comparison between our proposed method with VR experiment driving data. For safety, our driver model has a slightly higher collision rate compared to the VR experiment data. \par
\begin{table}[H]
\caption{Statistic results of our proposed approach compared with VR experiments}
\label{tab1}
\resizebox{\columnwidth}{!}{%
\begin{tabular}{@{}cccc@{}}
\toprule
\multicolumn{2}{c}{Method}                                                                                                                 & VR experiment & Ours             \\ \midrule
Safety                      & Collision rate                                                                                      & /             & \textbf{0.15\%}  \\ \cmidrule(lr){2-2}
\multirow{2}{*}{Efficiency} & Yielding rate                                                                                       & 51\%          & \textbf{75.03\%} \\
                                     & Average vehicle speed (m/s)                                                                          & 9.468         & \textbf{9.600}   \\ \cmidrule(lr){2-2}
\multirow{2}{*}{Smoothness} & Average vehicle jerk                                                                                & 0.897         & \textbf{0.843}   \\
                                     & \begin{tabular}[c]{@{}c@{}}Average maximum absolute \\ acceleration/deceleration (m/s\textsuperscript{2})\end{tabular} & 2.228         & \textbf{1.867}   \\ \bottomrule
\end{tabular}%
}
\end{table}
\vspace{-5pt}
To evaluate driving efficiency, the vehicle yielding rate and average vehicle speed are considered. The yielding rate observed in the simulation (75.03\%) is higher than in the VR experiment (51\%). This suggests that the algorithm is more cautious, which may slightly affect the average speed. However, the average speed in the simulation (9.600 m/s) is slightly faster compared to the experiment (9.468 m/s), indicating that our approach can maintain efficiency even with a greater yielding rate.  
Two parameters, jerk and maximum absolute acceleration/deceleration, are used to evaluate smoothness. The lower average jerk value and average maximum absolute acceleration/deceleration value in Table \ref{tab1} indicate that our proposed method can achieve smoother and more comfortable driving behavior.\par
In summary, the quantitative analysis shows that the proposed AV decision-making algorithm performs well in safety, efficiency, and smoothness. The similar average vehicle speed values between VR experimental data and simulation data indicate that our algorithm closely mimics real-world driving behavior. Additionally, the lower jerk values and maximum absolute acceleration/deceleration values in our simulations suggest that our method achieves smoother driving compared to the experimental data.\par

\section{Conclusion}
This paper proposes an innovative framework to address the decision-making challenges AVs face when interacting with pedestrians at the unsignalized intersection. First, we integrate the POMDP with behavioral game theory to model these interactions, capturing the uncertainty and dynamism of pedestrian behavior. Second, both the AV and pedestrian are modeled as DB-QCH models, accounting for human reasoning limitations and bounded rationality, thus enabling more realistic interaction simulations compared to traditional game theory approaches. Moreover, the dynamic updating mechanism for the opponent’s rationality degree is introduced, which allows the AV to adjust its strategies based on real-time observations. Finally, a trained neural network is developed to guide MCTS within the AV’s continuous action space, improving decision-making efficiency and effectiveness. \par
Simulation results demonstrate that our method excels in safety, efficiency, and smoothness, closely resembling real-world driving behavior. Although our model performs well, our current research is limited to a simple scenario of a single AV and a single pedestrian interaction. In the future, we will expand our scope to include the interaction between a single AV and multiple pedestrians, allowing the proposed AV decision-making algorithm to handle more complex scenarios.\par

\bibliographystyle{IEEEtran}

\begin{IEEEbiography}[{\includegraphics[width=1in,height=1.25in,clip,keepaspectratio]{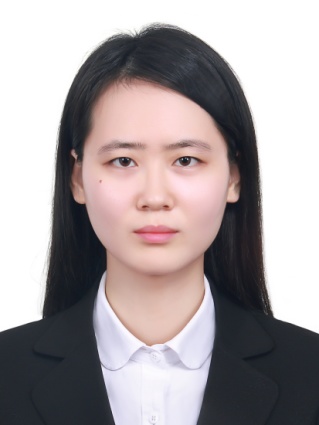}}]{Meiting Dang}received the B.S. and the M.S. degrees from Chang'an University, China, in 2017 and 2020, respectively. She is currently working toward the Ph.D. degree in James Watt School of Engineering with University of Glasgow, U.K. Her research interests include decision-making and planning of autonomous vehicles, autonomous vehicle-pedestrian interaction modeling based on game theory, and machine learning.
\end{IEEEbiography}

\begin{IEEEbiography}[{\includegraphics[width=1in,height=1.25in,clip,keepaspectratio]{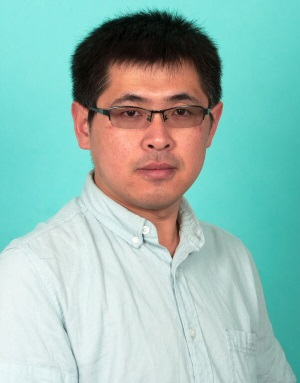}}]{Dezong Zhao} (Senior Member, IEEE) received the B.S. and M.S. degrees from Shandong University, Jinan, China, in 2003, and 2006, respectively, and the Ph.D. degree from Tsinghua University, Beijing, China, in 2010. His research interests include connected and autonomous vehicles, low carbon vehicles, machine learning and nonlinear control theory and applications. Dr. Zhao is a Fellow of the Higher Education Academy and was the recipient of the Excellence 100 Campaign of Loughborough University.
\end{IEEEbiography}

\begin{IEEEbiography}[{\includegraphics[width=1in,height=1.25in,clip,keepaspectratio]{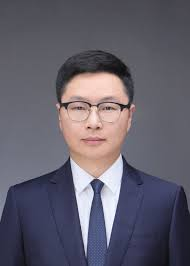}}]{Yafei Wang} (Member, IEEE) received the B.S. degree from Jilin University, Changchun, China, in 2005, the M.S. degree from Shanghai Jiao Tong University, Shanghai, China, in 2008, and the Ph.D. degree from The University of Tokyo, Tokyo, Japan, in 2013. He is currently a Professor of automotive engineering with the School of Mechanical Engineering, Shanghai Jiao Tong University. His research interests include state estimation and control for connected and automated vehicles.
\end{IEEEbiography}

\begin{IEEEbiography}[{\includegraphics[width=1in,height=1.25in,clip,keepaspectratio]{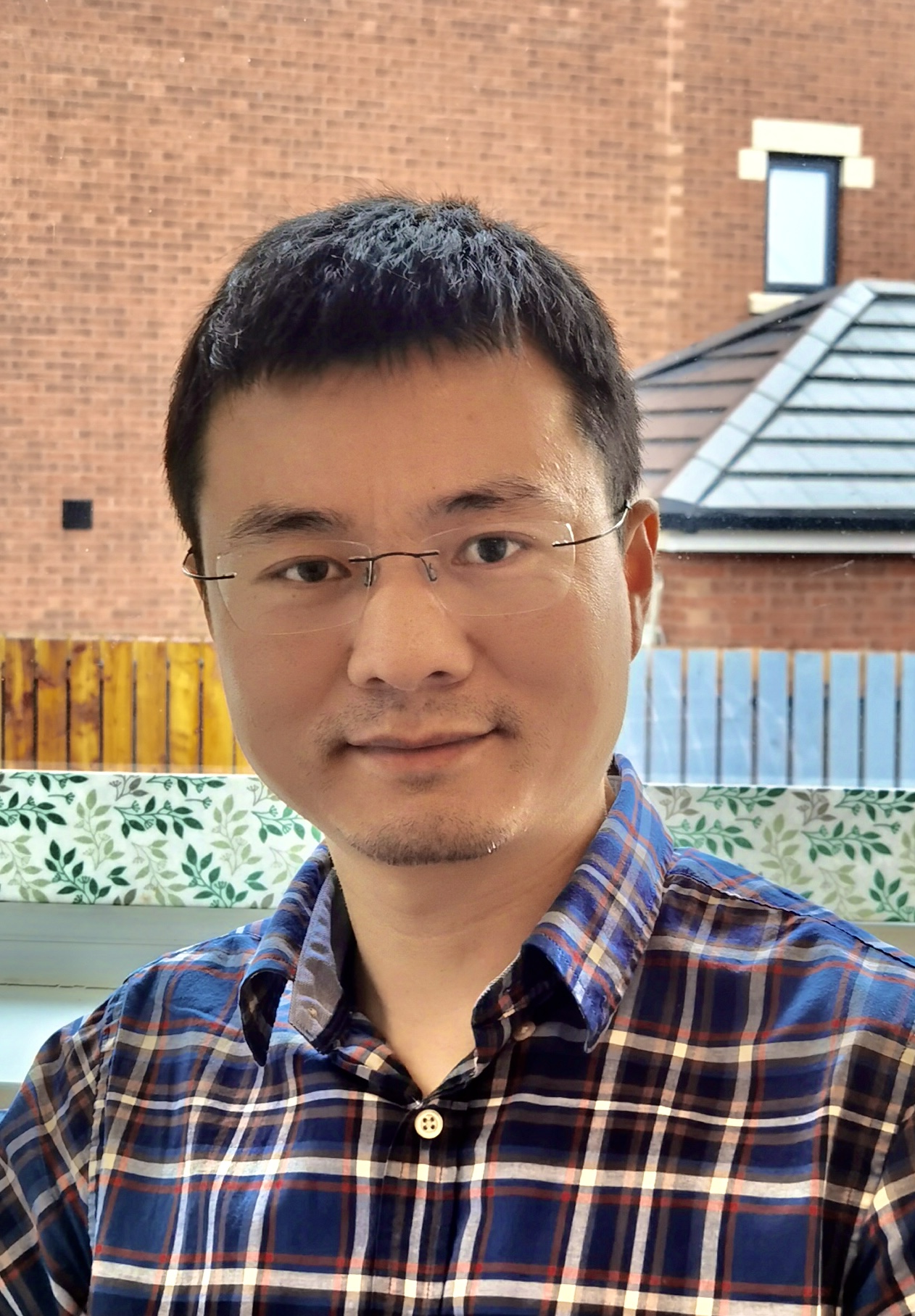}}]{Chongfeng Wei} (Member, IEEE) received his Ph.D. degree in mechanical engineering from the University of Birmingham in 2015. He is now a Senior Lecturer (Associate Professor) at University of Glasgow, UK. His current research interests include decision-making and control of intelligent vehicles, human-centric autonomous driving, cooperative automation, and dynamics and control of mechanical systems. He is also serving as an Associate Editor of IEEE TITS, IEEE TIV, IEEE TVT, and Frontier on Robotics and AI.
\end{IEEEbiography}

\end{document}